**Early Prediction of In-Hospital ICU Mortality Using Innovative First-Day Data: A Review**

Baozhu Huang[1#], Cheng Chen[1#], Xuanhe Hou[2#], Junmin Huang[3], Zihan Wei[3], Hongying Luo[3], Lu Chen[3], Yongzhi Xu[3,4], Hejiao Luo[5], Changqi Qin[3,4], Ziqian Bi[6], Junhao Song[7], Tianyang Wang[8], ChiaXin Liang[9], Zizhong Yu[1], Han Wang[3,4,10], Xiaotian Sun[11*], Junfeng Hao[3,4,12*], Chunjie Tian[1,4*]

[1]Department of Otorhinolaryngology, Affiliated Hospital of Guangdong Medical University, Zhanjiang 524000, Guangdong, China.

[2]Department of Radiation Oncology (MAASTRO), Maastricht University Medical Center, Maastricht, Netherlands, GROW School for Oncology and Reproduction, Maastricht University Medical Center, Maastricht, Netherlands.

[3]Key Laboratory of Prevention and Management of Chronic Kidney Disease of Zhanjiang City, Institute of Nephrology, Affiliated Hospital of Guangdong Medical University, Zhanjiang 524001, Guangdong, China.

[4]Guangdong Provincial Key Laboratory of Autophagy and Major Chronic Non-communicable Diseases.

[5]Department of Critical Care Medicine, Affiliated Hospital of Guangdong Medical University, Zhanjiang, China.

[6]Beijing University of Technology, Beijing, 100124, China.

[7]China Agricultural University, Beijing, 100083, China.

[8]Xi'an Jiaotong-Liverpool University, 111 Ren'ai Road, Suzhou, Jiangsu, 215123, China.

[9]JTB Technology Corp. No. 861, Siraya Blvd., Southern Taiwan Science Park, Shanhua Dist., Tainan City 741, Taiwan (R.O.C.).

[10] First clinical college, Guangdong Medical University, Zhanjaing, China

[11]Department of Gastroenterology, Second Medical Center, Chinese PLA General Hospital, Beijing 100853, China.

[12]Department of Family Medicine, Shengjing Hospital of China Medical University, Shenyang, 110022, Liaoning, China.

#These authors contributed equally.

*Corresponding authors. email: ygzhjf85@gmail.com, Han@hanziwww.me or tcjent@gmail.com, sky5150591@163.com

**Abstract**

**Background:** Critically ill patients admitted to the intensive care unit (ICU) often face a high risk of mortality. Early and accurate prediction of in-hospital mortality within 24 hours of ICU admission is crucial for facilitating clinical interventions, optimizing resource allocation, and improving patient outcomes. While traditional scoring systems have demonstrated utility, they often exhibit limited predictive accuracy and adaptability.

**Objective:** This review aims to systematically evaluate and benchmark innovative methodologies that leverage data available within 24 hours of ICU admission for predicting in-hospital mortality, with emphasis on advancements in machine learning, novel biomarker applications, and the integration of diverse data types.

**Methods:** A comprehensive literature search was conducted to identify relevant English-language studies. We analyzed studies that employed innovative techniques, such as advanced machine learning algorithms (e.g., ensemble methods, deep learning, and reinforcement learning), novel biomarkers (e.g., metabolomic and proteomic signatures), and multimodal data integration (e.g., structured Electronic Health Record (EHR) data with clinical notes). Key aspects, including study design, patient cohorts, data sources, features, methodologies, and performance metrics, were extracted and synthesized.

**Results:** Numerous studies demonstrated the superiority of innovative methods over traditional scoring systems, with machine learning models frequently achieving high discrimination (Area Under the Receiver Operating Characteristic curve values often exceeding 0.85 and sometimes >0.90). Large ICU databases, such as MIMIC-III/IV and eICU-CRD, were commonly used in these investigations. While novel biomarkers exhibited potential, their clinical applicability required further validation. The integration of unstructured data and the application of explainable AI (XAI) techniques emerged as notable trends, contributing to improved model performance and interpretability.

**Conclusion:** Innovative approaches, particularly those leveraging machine learning with initial 24-hour ICU data, hold significant potential for refining early mortality prediction. Future research should focus on external validation, clinical integration, and address challenges such as data heterogeneity and model interpretability to effectively facilitate the translation of these advancements into routine clinical practice.

# 1. Introduction

## 1.1. Challenge of ICU Mortality

Intensive Care Units (ICUs) are specialized hospital departments that provide comprehensive and continuous care for critically ill patients. Despite significant advancements in medical technology and critical care practices, mortality rates in ICUs remain a major concern worldwide. Studies have reported average ICU mortality rates ranging from 7% to 19%[1], with some analyses indicating figures between 9.3% and 26.2% in global audits, and even broader ranges, from 1% to 51%, particularly within elderly ICU patient populations[2]. This substantial mortality burden underscores the critical need for effective strategies to identify patients at highest risk of adverse outcomes.

Early and accurate risk stratification, especially within 24 hours of ICU admission, is paramount. This enables clinicians to make timely and informed decisions regarding treatment intensity, allocate resources efficiently, and engage in meaningful prognostic discussions with patients and their families[3]. The ability to predict mortality early can guide interventions aimed at mitigating risk and potentially improving survival rates.

## 1.2. Limitations of Traditional Scoring Systems

For decades, clinical practice has relied on traditional severity-of-illness scoring systems such as the Acute Physiology and Chronic Health Evaluation (APACHE), Simplified Acute Physiology Score (SAPS), and Sequential Organ Failure Assessment (SOFA)[4]. These systems typically use a predefined set of physiological variables, demographic data, and sometimes diagnostic information, usually collected within 24 hours of ICU admission, to generate a numerical score associated with mortality risk. While these scoring systems have been instrumental in benchmarking ICU performance, comparing patient populations in research, and delivering general assessments of illness severity, their utility in predicting individual patient outcomes remains limited. Their predictive accuracy, commonly measured by the Area Under the Receiver Operating Characteristic (AUROC) curve, generally ranges between 0.70 and 0.79. This moderate performance can be attributed to several factors, including their dependence on linear models, insufficient calibration to local patient demographics or evolving medical practices, and an inherent failure to capture the full complexity and dynamic interactions underlying critical illness[5]. Furthermore, these systems exhibit limited ability to harness the vast amounts of high-resolution data now routinely obtained through modern monitoring and electronic health records (EHRs).

## 1.3. Promise of Innovative Methods

The widespread adoption of EHRs and the availability of comprehensive critical care databases, such as the Medical Information Mart for Intensive Care (MIMIC-III/IV)[5] and the eICU Collaborative Research Database (eICU-CRD)[6], have transformed predictive modeling in intensive care medicine.

These resources provide unprecedented opportunities to develop and validate more sophisticated and accurate tools for ICU prognostication. Over the past few years, there has been an increase in innovative methods that leverage these comprehensive data sources. These include advanced machine learning (ML) algorithms capable of identifying subtle and complex patterns indicative of risk, the discovery and application of novel molecular biomarkers (e.g., from metabolomics, proteomics), and advanced techniques for integrating diverse data modalities, such as unstructured clinical documentation. A significant focus of this research involves leveraging data collected within the first 24 hours of ICU admission, a critical timeframe during which early interventions exert the most substantial impact on patient outcomes[3].

This review aims to provide a comprehensive overview and critical analysis of recent English-language studies that employ innovative methods using first-day ICU data to predict in-hospital mortality. We explored the various methodologies, the types of data being utilized, and the reported predictive performance of these novel approaches. A key component of this review involves the establishment of standardized evaluation metrics, designed to offer a structured comparison of these studies, thereby highlighting current trends, identifying successful strategies, and pointing towards areas for future research and development in this rapidly evolving field. Ultimately, this review provides both researchers and clinicians with insights into evolving ICU mortality prediction methodologies and their potential to enhance patient care.

## 2. Innovative Methodologies for Early ICU Mortality Prediction

To enhance prognostic accuracy in critical care, researchers are pioneering methods that leverage data collected within 24 hours of admission, moving beyond the limitations of traditional scoring systems.

### 2.1. Advanced Machine Learning (ML) Algorithms

Machine learning has emerged as a transformative force in medical informatics, particularly in developing predictive models. Unlike traditional statistical models, which often rely on predefined assumptions and linear relationships, ML algorithms can learn complex, non-linear patterns directly from data[3]. The initial 24-hour period following ICU admission generates a wealth of data, including vital signs, laboratory results, demographic information, and clinical interventions, which serve as rich input for these models.

#### 2.1.1. Ensemble Methods

Ensemble methods, which integrate predictions from multiple base learners to improve overall performance and robustness, have shown promise in critical care prognostication.

**Gradient Boosting Machine (GBM):** Advanced gradient boosting algorithms, particularly XGBoost and LightGBM, have become widely adopted in clinical predictive modeling. Johnson & Mark (2018) demonstrated that a Gradient Boosting model trained on data from the first 24 hours achieved a significantly higher AUROC (0.927) compared to SAPS II (0.809)[7]. This underscores the potential of ML to extract more predictive information from the same data. Pang et al. (2022) further illustrated this advantage using XGBoost, incorporating APACHE III and LODS subscores, and achieved an AUROC of 0.918, outperforming Logistic Regression (LR), Support Vector Machines (SVM), and Decision Trees[8]. This approach effectively integrates the clinical knowledge encoded in established scoring systems with the superior pattern recognition capabilities of ML. Similarly, Liao et al. (2024) reported XGBoost's superior performance in predicting mortality in acute pancreatitis-associated acute kidney injury (AP-AKI) patients[9]. Consistently, Tu et al. (2023) reported an AUROC of 0.921 for LightGBM in traumatic brain injury (TBI) patients using first-day ICU features[10].

**Random Forest (RF):** This ensemble technique, which builds multiple decision trees and merges their outputs, has also demonstrated effectiveness. Safaei et al. (2022) used RF as a strong baseline model in their study, which ultimately highlighted CatBoost[6]. Villar et al. (2025) found that ML methods, including RF, when applied to data from the first 24 hours in patients with Acute Hypoxemic Respiratory Failure (AHRF), achieved an AUROC of 0.88, comparable to a logistic regression model enhanced with a genetic algorithm for feature selection[9].

**Stacked Ensemble:** Ren et al. (2022) developed a stacked ensemble model that integrated conventional clinical severity scoring results (APACHE scores) with predictions from base learners like LR, Naive Bayes, RF, and XGBoost. This hierarchical approach yielded an AUROC of 0.879, demonstrating that combining diverse models can lead to improved performance[11].

**CatBoost:** Safaei et al. (2022) introduced E-CatBoost, an optimized model using only the top ten most informative features from the first 24 hours. This model achieved a notable AUROC of 0.91 for the entire patient population within the eICU-CRD, outperforming traditional scoring systems like APACHE IVa[6]. This highlights that sophisticated feature selection combined with powerful algorithms can yield highly accurate yet parsimonious models.

The consistent success of ensemble methods suggests their suitability for the complex and frequently noisy data encountered in the ICU. By aggregating the "wisdom" of multiple models, they tend to exhibit greater robustness and reduced susceptibility to overfitting compared to individual models. Furthermore, the strategy of using components of established clinical scores (like APACHE or LODS subscores) as input features for these advanced ML models represents a synergistic approach. It allows these models to leverage the distilled physiological knowledge embedded within these scores while applying more powerful analytical techniques to uncover deeper patterns.

#### 2.1.2. Deep Learning Approaches

Deep learning, a subfield of ML based on artificial neural networks with multiple layers (deep architectures), has shown remarkable success across various domains, including healthcare. These models can automatically learn hierarchical feature representations from raw data, potentially reducing the need for manual feature engineering. Liu et al. (2024) developed a deep learning neural network model to predict ICU mortality in mechanically ventilated patients using only 12 variables from the MIMIC-III database, including mechanical ventilation duration. This model demonstrated a 7.06% improvement in AUROC (0.879) over existing literature, showcasing the efficiency of deep learning in feature representation[12]. Wernly et al. (2023) employed a deep learning model that integrated embedded text from multiple data sources with Recurrent Neural Networks (RNNs) to predict unplanned ICU transfers and in-hospital deaths. With a 24-hour assessment window and a 1-day prediction horizon, their model achieved an AUROC of 0.905, demonstrating the capability of deep learning to integrate heterogeneous data, including unstructured text[13]. Li et al. (2022) found that a Neural Network (NN) model yielded the best predictive performance (AUROC 0.764 in the test cohort and 0.674 in external validation) for 28-day mortality in ICU patients with heart failure and hypertension, using initial 24-hour data from MIMIC-IV and eICU-CR[14]. Deep learning models exhibit good performance in managing the temporal characteristics of ICU data (e.g., time-series vital signs) and integrating diverse data types. Their ability to learn representations directly from data can uncover patterns that might be missed by traditional methods or require extensive manual feature engineering. However, these models typically require large datasets for training to avoid overfitting and can be computationally demanding. Furthermore, their inherent "black box" nature poses challenges for clinical interpretability, although this is being addressed by the application of XAI techniques.

#### 2.1.3. Novel ML Paradigms

Beyond established supervised learning techniques, significant emphasis has been placed on the exploration of more advanced paradigms:

**Temporal Difference (TD) Learning:** Frost et al. (2024) introduced a TD learning framework, a concept from reinforcement learning, for real-time ICU mortality prediction[1]. This approach models patient trajectories as sequences of states and learns to predict outcomes based on transitions between these states, rather than relying solely on a single snapshot in time. While their model used a 7-day lookback window for feature construction rather than strictly initial 24-hour data for each prediction point, its continuous and adaptive nature is innovative. The aim is to improve model robustness, which was demonstrated through training on MIMIC-IV and validation on an external dataset (Salzburg Intensive Care dataset)[1]. This dynamic risk assessment capability could be highly beneficial in the ICU, where patient conditions can evolve rapidly.

**Causal Inference-Guided Deep Learning:** Wang et al. (2025) proposed the Causal Relationship Informed Superior Prediction (CRISP) framework[2], which integrates a causal graph into a Transformer-based deep learning architecture to predict mortality using data from the first 48 hours. It also features a Counterfactual Minority Generation Module (CMG) to address class imbalance by generating a synthetic minority class guided by causal relationships. The model achieved high AUROCs (0.9480 on MIMIC-III, 0.9171 on MIMIC-IV). This approach signifies a move beyond purely correlational models towards understanding the underlying causal mechanisms of mortality, which could ultimately lead to more interpretable and actionable predictions.

These novel paradigms, while still in the early stages of exploration for ICU mortality prediction, represent exciting future directions. TD learning offers a more dynamic and potentially robust way to model patient risk over time. Causal inference aims to provide a deeper understanding and potentially guide interventions, shifting AI from a purely predictive tool to a more comprehensive decision support system.

## 2.2. Novel Biomarker Discovery and Application

The identification of novel biomarkers from easily accessible biological samples collected early in ICU admission opens new possibilities for improving mortality prediction. These biomarkers can reflect underlying pathophysiological processes that may not be immediately apparent from standard clinical data.

### 2.2.1. Metabolomics

Metabolomics involves the comprehensive analysis of small-molecule metabolites within biological systems. Silva et al. (2024) explored the use of serum metabolomic fingerprints obtained via Fourier-Transform Infrared (FTIR) spectroscopy in COVID-19 ICU patients[15]. Their study identified significant spectral differences between survivors and non-survivors. By applying a Fast-Correlation-Based Filter (FCBF) for feature selection, Naïve Bayes models achieved an AUROC of 0.79 for predicting mortality within the first 48 hours of ICU admission. Accordingly, FTIR spectroscopy represents a rapid, economical, and minimally invasive technique in critical care settings.

While not directly predicting mortality, Huang et al. (2025) identified a panel of three serum metabolites (inosine, creatine, 3-hydroxybutyric acid) from first-day samples that could diagnose Sepsis-Associated Acute Kidney Injury (SA-AKI) with an AUROC of 0.90[16]. Importantly, their IC3 model score correlated with established severity scores like SOFA and APACHE II, which are themselves predictors of mortality. This suggests an indirect prognostic value for these metabolites in assessing overall critical illness severity. Metabolomic approaches provide a snapshot of the patient's current physiological state. The dynamic changes in metabolite profiles can offer early indications of deterioration. While promising, the translation of these findings into routine clinical tools requires further validation in larger, more diverse cohorts and the standardization of analytical methods.

#### 2.2.2. Proteomics and Genomics

Proteomic and genomic/transcriptomic profiling enables the identification of molecular signatures associated with disease severity and outcome. A review by Chen et al. (2023) highlighted several studies using proteomics and transcriptomics (e.g., microRNAs or miRNAs) for outcome prediction in ICU patients[17]. For example, Ruiz-Sanmartín et al. (2022) identified a 22-protein signature that predicted 30-day mortality in septic shock patients with an AUROC of 0.81[18]. Molinero et al. (2022) developed a model based on five proteins (AGR2, NQO2, IL-1α, OSM, and TRAIL) for predicting in-ICU mortality in septic patients[19]. Dong et al. (2021) identified that plasma Insulin-like Growth Factor Binding Protein 7 (IGFBP7) as a significant predictor of 28-day mortality in ARDS patients. Liu et al. (2017) reported that interleukin-10 concentration could predict ICU mortality with a sensitivity of 73.3% and a specificity of 90.5[20]. Taken together, these molecular markers provide insights into specific pathological pathways, such as inflammation, immune response, or organ damage, offering a more granular understanding of a patient's risk. However, similar to metabolomics, the clinical utility of proteomic and genomic markers depends on the development of reliable, rapid, and cost-effective assays, as well as extensive validation.

### 2.3. Integration of Multi-modal and Unstructured Data

Modern ICUs generate a vast amount of data beyond structured numerical values. Recent methodological advances have emphasized the integration of diverse data sources, particularly unstructured clinical text, to build more comprehensive predictive models. Clinical documentation, such as physician admission notes, progress reports, and nursing narratives, contains a wealth of information regarding patient history, clinical decision-making, and subtle observations that are often not captured in structured data fields. Chiu et al. (2023) demonstrated the value of this approach by combining eight structured variables (vital signs, GCS, age) from the first 24 hours with features derived from unstructured initial physician diagnoses using Latent Dirichlet Allocation (LDA) topic modeling[21]. This integration significantly improved mortality prediction in MIMIC-III patients, with the AUROC for 3-day mortality increasing from approximately 0.83 (using structured data alone) to 0.8820 (using combined data). Wernly et al. (2023) also achieved high predictive performance (AUROC 0.905 for 1-day mortality prediction) using a deep learning model that incorporated embedded text from multiple data sources, including clinical notes, processed by RNNs[22]. An earlier study by Ghassemi et al. found that adding topic-modeled features extracted from the initial 24-hour clinical notes to a classifier based on SAPS I scores improved the AUROC from 0.78 to 0.82[23]. These studies collectively demonstrate that unstructured clinical text is a rich yet often underutilized source of predictive information. Natural Language Processing (NLP) techniques are crucial tools for leveraging this data, enabling predictive models to gain a more comprehensive understanding of patient status and thereby enhance prognostic accuracy. The challenge lies in the complexity of NLP implementations and ensuring the robust and secure processing of sensitive narrative data.

### 2.4. Explainable AI (XAI) and Causal Discovery

As ML models, particularly deep learning architectures, become more complex and achieve higher accuracy, their "black box" nature can pose a significant barrier to clinical adoption. For effective implementation in healthcare settings, clinicians need to understand the rationale behind a model's prediction, enabling both trust and informed clinical decision-making.

**Importance of Interpretability:** Over the past few years, the demand for transparency in AI-driven medical tools has been growing[24]. XAI aims to make the decision-making process of these models comprehensible to human understanding. Several studies reviewed have incorporated XAI techniques, predominantly SHapley Additive exPlanations (SHAP), to identify which input features contribute most significantly to the model's predictions. Ren et al. (2022) used SHAP to interpret their stacked ensemble model, confirming the importance of APACHE-related features[11]. Liu et al. (2024) employed SHAP analysis with their deep learning model for mechanically ventilated patients, identifying respiratory failure as a key predictor[9]. Li et al. (2022) aimed to enhance the interpretability of their neural network model for heart failure patients using SHAP[14]. Safaei et al. (2022) used SHAP to identify critical cross-disease features, including age, heart rate, respiratory rate, BUN, and creatinine in their E-CatBoost model[6]. Similarly, Tu et al. (2023) used SHAP to identify influential features for TBI mortality prediction in their LightGBM model, including vasopressor use and GCS components[10]. The adoption of XAI methods represents a positive development, as it helps bridge the gap between a model's statistical performance and its clinical utility. By providing insights into feature importance and, in some cases, patient-specific explanations, XAI can increase clinician confidence and facilitate the identification of potentially novel risk factors. Moving beyond correlation to understand causation represents a more ambitious yet potentially more impactful goal. Causal discovery aims to identify the underlying mechanisms and cause-effect relationships within the data.

The CRISP framework developed by Wang et al. (2025) is a notable example that explicitly incorporates a causal graph to guide the deep learning model and harnesses counterfactual reasoning for data augmentation[2]. This approach seeks not only to predict mortality but also to understand how different factors contribute to that risk, potentially offering insights into modifiable factors. Bravi et al. (2021) used Dynamic Bayesian Networks (DBNs) to model the interrelationships and temporal progression of organ failures in ICU patients, collecting data at admission, on day 2, and on day 7[25]. While not exclusively focused on first-day mortality prediction, DBNs are inherently suited for representing conditional dependencies that, under certain assumptions, can be interpreted causally, thus offering a means to understand the dynamic pathways leading to adverse outcomes. The shift towards incorporating causal inference represents a significant step. If models can reliably identify not only that a patient is at high risk, but also the underlying reasons, and potentially simulate the impact of interventions on modifiable risk factors, they transition from passive predictors to active decision-support tools. This represents a crucial advancement for personalized medicine in the ICU.

## 3. Benchmark Analysis of Predictive Models

To facilitate a clear comparative overview of the diverse innovative approaches discussed, this section introduces a benchmark table that summarizes key studies focusing on predicting ICU mortality using data from the first 24 hours of admission. This table is designed to consolidate information on methodologies, data sources, patient cohorts, key features, performance metrics, and principal findings, facilitating an understanding of the current state-of-the-art and identifying common trends and challenges. Analyzing the studies presented in Table 1 revealed several key trends and considerations in the field of early ICU mortality prediction.

(1) Model Performance. Empirical evidence demonstrates the superior performance of ML models, particularly ensemble methods like Gradient Boosting (XGBoost, LightGBM, CatBoost) and deep learning approaches, over traditional scoring systems when using initial 24-hour ICU data[26]. AUROC values frequently reported in the range of 0.85 to above 0.90 signify a substantial improvement in discriminatory power. For example, Johnson & Mark (2018) reported a Gradient Boosting model which outperformed SAPS II (AUROC 0.927 vs. 0.809) using initial 24-hour ICU data[4]. This suggests that the inherent ability of these algorithms to capture complex, non-linear interactions within the rich, high-dimensional data available early in an ICU stay provides a distinct advantage. This capacity is crucial due to the frequently non-linear nature of physiological responses in critically ill patients, which often involve complex interdependencies that simpler models may fail to adequately capture.

(2) Data Sources and Generalizability. Most studies leverage large, publicly available multicenter databases such as MIMIC-III, MIMIC-IV, and eICU-CRD. These databases provide extensive data for model development and internal validation. However, external validation on independent datasets from different geographical locations or healthcare systems constitutes a critical aspect for clinical translation. Studies like Frost et al. (2024), which validated their TD learning model on the Salzburg Intensive Care dataset, Liao et al. (2024) using Xiangya Hospital data[9], Li et al. (2022) using eICU-CRD as external validation for a MIMIC-IV trained model[14], and Wang et al. (2025) using WCHSU data[2], are crucial. These efforts address the concern that models might exhibit overfitting to the characteristics of a single dataset or region, thereby lacking broad applicability. The significant variability observed in reported ICU mortality rates across different settings further emphasizes the need for models that are robust and generalizable.

(3) Feature Sets. A broad spectrum of features is leveraged across these models. Some models achieve high performance with a relatively limited, yet carefully curated set of readily available clinical variables.

For instance, Liu et al. (2024) used only 12 variables in their deep learning model, and Safaei et al. (2022) developed E-CatBoost using just ten features[6]. This demonstrates that effective feature selection or advanced model design can yield high predictive accuracy without necessitating an exhaustive list of inputs, which offers practical benefits for implementation. Conversely, other studies explore the integration of novel data types, such as metabolomic fingerprints[3], specific protein panels[27], or features derived from unstructured clinical notes[21]. These approaches aim to capture deeper physiological insights or contextual information that might be missed by conventional structured data alone. The selection of features often reflects a trade-off between the potential for increased accuracy and the practicality of data acquisition in a standard clinical workflow.

(4) Innovation Spectrum. The reviewed studies demonstrate a spectrum of innovation. Some studies build upon well-established ML techniques, adapting and optimizing them for the ICU context (e.g., stacked ensembles[11], specialized CatBoost models[6]). Conversely, others introduce more paradigm-shifting approaches, such as Temporal Difference learning for dynamic risk assessment or the integration of causal inference into deep learning frameworks. The development of models using novel biomarkers, such as metabolomic profiles, also represents a significant innovative direction. This diversity in approaches is beneficial for the advancement of the field, pushing the boundaries of what is possible in early prognostication. The ultimate clinical value of these innovations will depend not only on their predictive power but also on their interpretability, ease of integration into clinical workflows, and cost-effectiveness.

## 4. Interpretability and the Path to Clinical Trust

The field of ICU mortality prediction using initial 24-hour data is rapidly evolving, driven by advancements in computational power, data availability, and novel analytical techniques. This review highlights a clear trend towards more sophisticated, data-intensive methods that consistently outperform traditional approaches.

The primary strength of the innovative methods discussed, particularly those based on machine learning, lies in their enhanced predictive accuracy[4]. By leveraging complex algorithms, these models can identify subtle patterns and interactions within the first 24 hours of ICU data that are often missed by conventional scoring systems. This allows for a more comprehensive and individualized risk assessment. Furthermore, the ability to integrate diverse data types, including continuous physiological monitoring, comprehensive laboratory panels, unstructured clinical notes[21], and novel molecular biomarkers, offers a more comprehensive view of the patient's condition. Emerging paradigms, such as temporal difference learning, also hold promise for dynamic risk assessment, adapting to the patient's evolving status, which is highly significant in the critical care setting. However, these innovative approaches are not without their challenges. A significant concern, especially with complex models like deep neural networks, is their "black box" nature, which can hinder clinical adoption. Clinicians may be reluctant to rely on predictions without understanding their basis. This issue is being actively addressed by the burgeoning field of XAI. Another major hurdle is the need for large, high-quality datasets to train robust models. Issues such as missing data[11], which is common in real-world clinical settings, and class imbalance (where mortality is a less frequent outcome), necessitate careful methodological handling. The risks of overfitting to training data and ensuring generalizability to new, unseen patient populations and different healthcare environments remain critical challenges, underscoring the necessity of rigorous external validation. Finally, the practical implementation of these models into busy clinical workflows, along with the potential cost and accessibility of novel biomarker assays, presents further barriers to widespread adoption.

The definition and scope of "data" used for ICU prognostication are expanding significantly. Traditionally, prediction models have relied on a limited set of structured physiological and demographic variables. However, innovative methods are now demonstrating the capacity to extract valuable predictive signals from previously underutilized or inaccessible data sources. The successful integration of unstructured clinical narratives through NLP and topic modeling techniques[14] allows for the incorporation of rich contextual information, physician insights, and nuanced patient history that structured data often fails to capture. Similarly, the exploration of omics data, such as metabolomic and proteomic[16] profiles, provides a window into the molecular underpinnings of critical illness. This ability to harness a broader spectrum of information available from day one is a key driver behind the improved performance of these newer models. The challenge lies in standardizing these diverse data types and developing robust methods for their combined analysis.

For predictive tools to be effectively integrated into clinical practice, clinician trust remains essential. Even highly accurate models may fail if their logic is unclear. Importantly, XAI bridges this gap. Techniques such as SHAP and LIME are increasingly being used to provide insights into which features drive a model's prediction for an individual patient. This enhanced transparency can help clinicians assess whether the model's reasoning aligns with their clinical knowledge, identify potentially overlooked factors, and ultimately foster confidence in the technology. Beyond interpretability, the increasing adoption of causal inference in predictive modeling, as exemplified by the CRISP framework[2], represents a significant advancement. While standard ML models identify correlations, causal models aim to understand the underlying cause-and-effect relationships. If a model can not only predict a high risk of mortality but also identify modifiable factors contributing to that risk, it transforms from a passive prognostic tool into an active decision support system. This could empower clinicians to implement more targeted interventions. For example, understanding that a specific, modifiable physiological parameter is a strong causal driver of predicted mortality for a patient could prompt a more aggressive therapeutic approach directed at that parameter.

Several methodological challenges are consistently encountered in developing ICU mortality prediction models. Class imbalance, characterized by a significantly smaller number of deceased patients compared to survivors, is a prevalent issue[11]. If not addressed, models can become biased towards predicting the majority class (survival), resulting in poor sensitivity for mortality prediction. Techniques such as Synthetic Minority Over-sampling Technique (SMOTE[11]), using weighted loss functions, or more advanced methods like the Counterfactual Minority Generation[28] (CMG) module within the CRISP framework are employed to mitigate this problem. Missing data constitutes another pervasive issue in real-world clinical datasets. ICU data collection can be intermittent or incomplete for various reasons. Robust imputation techniques or models inherently designed to handle missing values are crucial for developing practical and reliable predictive tools. The high dimensionality of modern ICU datasets, especially when incorporating -omics data or extensive EHR variables, also requires careful feature selection and dimensionality reduction strategies to prevent overfitting and improve model efficiency. Addressing these common data challenges with methodological rigor is crucial for developing models that are not only accurate during development but also perform reliably in diverse clinical settings.

## 5. Limitations of Current Research and Future Directions

Despite the promising results, contemporary research on innovative ICU mortality prediction reveals several limitations that point to future research directions.

(1) Prospective Validation: Most studies are retrospective, relying on existing datasets. Prospective studies are critically needed to validate these models in real-time clinical settings and assess their actual impact on clinical decision-making and patient outcomes.

(2) Standardization and Comparability: There is a need for greater standardization in data collection, feature definition, and the reporting of performance metrics to facilitate more robust comparisons across different studies and methodologies.

(3) Clinical Integration and Usability: Significant research is required to understand how to best integrate these predictive tools into existing clinical workflows. This includes developing user-friendly interfaces, ensuring timely delivery of predictions, and providing clear guidance on the interpretation and application of model outputs without causing alert fatigue or excessive clinician burden.

(4) Beyond Mortality Prediction: While mortality represents a crucial endpoint, innovative methods using initial 24-hour data could also be applied to predict other important ICU outcomes, such as length of stay, development of specific complications (e.g., sepsis, ARDS, AKI), or response to specific therapies. Some studies are already moving in this direction[13].

(5) Ethical Considerations: As AI models become more sophisticated and integrated into healthcare

systems, it is imperative to address the ethical implications, including potential biases in algorithms (e.g., based on race, socioeconomic status), fairness in resource allocation decisions informed by these models, and accountability for model predictions and their consequences.

(6) Hybrid Models: Future research could explore the development of hybrid models that combine the strengths of different innovative approaches, such as integrating biomarker data with advanced ML algorithms that also process clinical notes.

(7) Causal Inference for Intervention: Further development of causal inference models is warranted to move beyond prediction towards identifying modifiable risk factors and suggesting personalized interventions that could improve patient outcomes.

**6. Conclusion**

The prediction of in-hospital mortality using data from the first 24 hours of ICU admission has undergone a significant transformation with the application of innovative methodologies. Machine learning algorithms, including sophisticated ensemble techniques and deep learning models, have consistently demonstrated their ability to surpass the predictive accuracy of traditional scoring systems. This improvement is largely attributable to their capacity to analyze complex, high-dimensional data and identify subtle patterns that are often missed by conventional methods. The exploration of novel biomarkers, derived from metabolomic and proteomic analyses, presents a promising avenue for identifying early molecular changes indicative of severe illness and a poor prognosis. Concurrently, the integration of previously underutilized unstructured data, such as clinical notes, through advanced natural language processing techniques, is enriching the predictive landscape by incorporating valuable contextual information. Despite these advancements, the journey from a high-performing algorithm in a research setting to a trusted and effective tool in routine clinical practice is fraught with challenges. The "black box" nature of some complex models necessitates the continued development and application of XAI techniques to foster clinical trust and understanding. Furthermore, the critical need for rigorous external validation across diverse patient populations and healthcare systems cannot be overstated to ensure model robustness and generalizability. Addressing methodological issues such as data heterogeneity, missing values, and class imbalance remains paramount.

Future research should focus on conducting prospective validation studies, developing strategies for seamless and ethical integration of these tools into clinical workflows, and expanding their application beyond mortality to other critical ICU outcomes. The progression towards causal inference models holds the potential to transform these tools from mere predictors to guides for targeted interventions. Ultimately, the continued synergistic collaboration between data scientists, clinicians, and biomedical researchers will be essential to harness the full potential of these innovative approaches, with the overarching goal of personalizing critical care and improving outcomes for the most vulnerable patients.

**Table 1. Benchmark of Studies Predicting ICU Mortality Using First-Day Data.**

| Study (Author, Year, Ref.) | Patient Cohort (Size, Type) | Data Source(s) | Key First-Day Features/Variables Used | Innovative Method(s) Employed | Primary Outcome Predicted | Key Performance Metrics (Test/Validation Set) | Salient Conclusions/Findings Related to First-Day Prediction |
|---|---|---|---|---|---|---|---|
| **Silva et al., 2024** [3] | 44 COVID-19 ICU patients (21 discharged, 23 deceased) | Single-center, prospective | Serum metabolomic fingerprint (FTIR spectra) | FTIR spectroscopy, Fast-Correlation-Based Filter (FCBF), Naïve Bayes | In-hospital mortality | AUC: 0.79 (first 48h) | FTIR spectroscopy shows promise as a rapid and economical tool for mortality prediction in COVID-19 ICU patients using early serum samples. |
| **Ren et al., 2022** [12] | 91,713 ICU encounters (general) | MIMIC-III (via WiDS Challenge 2020) | Demographics, vitals, labs, APACHE scores (186 features initially) | Stacked Ensemble Model (SEM) integrating APACHE scores with LR, NB, RF, XGBoost; SHAP for interpretability | ICU mortality | AUC: 0.879 (SEM with SetS features) | Stacked ensemble model incorporating severity scores outperforms models without this integration. APACHE features are highly important. |
| **Johnson & Mark, 2018** [7] | 50,488 adult ICU stays (general) | MIMIC-III v1.4 | Physiology, labs, demographics (first 24h) | Gradient Boosting (GB), Logistic Regression (LR) | In-hospital mortality | GB: AUROC = 0.927 [0.925, 0.929]; LR: AUROC = 0.896 [0.892, 0.899] | GB model greatly outperformed SAPS II (AUROC = 0.809). ML with initial 24h data significantly |
| | | | | | | | improves prediction. |
| **Liu et al., 2024** [14] | 16,499 mechanically ventilated ICU patients | MIMIC-III | 12 variables: Age, SAPS II, respiratory failure, CHF, diabetes, Hb, lactate, BUN, PaO2, PaCO2, vent duration (first 24h data for most) | Deep Learning Neural Network; SHAP analysis | ICU mortality | AUROC: 0.879 [0.861-0.896], Accuracy: 0.859 | Proposed DL model outperformed other ML models and existing literature with fewer features. Respiratory failure and ventilation duration were significant predictors. |
| **Frost et al., 2024** [1] | >65,000 patients (undifferentiated ICU) | MIMIC-IV, Salzburg Intensive Care dataset (SICdb) | Time-series data (up to 7-day lookback for each state, including the first day) with 396 measurements + age, gender, weight | Temporal Difference (TD) Learning, CNN-LSTM base model | Inpatient mortality (various horizons) | AUROC for TD models generally outperformed supervised learning baselines, especially on external validation (specific first-day only AUROC not isolated) | TD learning improves model robustness for long-term mortality prediction, maintained on external validation. |
| **Wernly et al., 2023** [22] | Development: 682,041 patients, Test: 170,579 patients (general hospital, including | Single large academic medical center EHR | Embedded text from multiple data sources (clinical notes, lab reports) & structured data (assessed at 24h for 1-day | Deep Learning (RNNs) | Composite: Unplanned ICU transfer or In-hospital death | AUROC: 0.905 [0.901–0.908] (for 24h assessment, 1-day prediction window for | Deep learning with embedded text from various sources effectively predicts adverse outcomes, including death. |
| | ICU transfers) | | prediction) | | | death/ICU transfer) | |
| **Pang et al., 2022** [8] | 14,110 ICU patients (general) | MIMIC-IV | Subscores of LODS and physiology subscores of APACHE III (APS III) from the first 24h | XGBoost, LR, SVM, Decision Tree | ICU mortality | XGBoost: AUROC = 0.918 [0.915–0.922] | XGBoost using APACHE III and LODS subscores performed best. GCS, respiratory rate score, and acid-base score were the most important factors. |
| **Liao et al., 2024** [15] | Training: 1089 AP-AKI patients; External Validation: 176 AP-AKI patients | MIMIC-IV, eICU-CRD, Xiangya Hospital data | Demographics, labs, vitals, comorbidities, treatment (first 24h) | XGBoost, RF, LR | In-hospital mortality | XGBoost Training AUROC: 0.89 [0.86-0.92]; XGBoost External Validation AUROC: 0.83 [0.77-0.90] | XGBoost outperforms LR and RF in predicting mortality in AP-AKI patients, demonstrating good performance in external validation. |
| **Li et al., 2022** [16] | MIMIC-IV: 3,458 patients; eICU-CRD: 6,582 patients (HF with hypertension) | MIMIC-IV, eICU-CRD | 22 clinical parameters (first 24h) | Neural Networks (NN), LR, SVM, RF; SHAP for interpretability | 28-day mortality | NN Test AUROC (MIMIC-IV): 0.764; NN External Validation AUROC (eICU): 0.674 | NN model performed best. SHAP aided interpretability. |
| **Villar et al., 2025** [17] | 1193 AHRF patients | Multicenter Spanish ICU cohort | Demographics, comorbidities, APACHE II, SOFA, organ failures, ventilator settings, gas | Logistic Regression with Genetic Algorithm, MLP, RF, SVM | ICU death | AUROC: 0.88 [0.86–0.90] (ML methods); Validation AUROC: 0.83 [0.78– | ML and traditional methods with feature selection achieved good prediction for AHRF mortality |
| | | | exchange (at diagnosis and within 24h) | | | 0.88] | using 24h data. Six key predictors identified. |
| **Yeh et al., 2024** [5] | 79,657 admissions (general ICU) | MIMIC-IV, CORE (Taipei hospital) | ADM model: 18 features (demographics, premorbid, GCS, vitals upon admission). 24H model: 40 features (adds diagnoses, treatments, labs). | Logistic Regression (LR), Gradient Boosted Trees (GBT), Deep Learning (DL) | ICU mortality | ADM GBT (Testing): AUROC 0.856, AUPRC 0.331. 24H GBT (Testing): AUROC 0.910, AUPRC 0.473 | GBT models performed best for both upon-admission and 24h predictions. Early prediction is feasible and valuable. |
| **Safaei et al., 2022** [10] | >200,000 ICU admissions (divided into disease groups) | eICU-CRD v2.0 | 29 numerical & 26 categorical features (first 24h); E-CatBoost used top 10 | CatBoost, E-CatBoost (optimized CatBoost); SHAP, LIME for XAI | ICU discharge mortality | CatBoost (overall): AUROC 0.91 [std:0.004]; E-CatBoost (overall): AUROC 0.87 [std:0.004] | CatBoost and E-CatBoost outperformed traditional scores and other ML models. XAI identified key cross-disease predictors. |
| **Chiu et al., 2023** [6] | 27,550 adult ICU patients | MIMIC-III | 8 structured variables (vitals, GCS, age) + unstructured initial diagnosis text (first 24h) | LDA for topic modeling; AdaBoost, Bagging, Gradient Boosting, LightGBM, LR, MLP, SVC, XGBoost | Mortality at 3, 30, 365 days | Gradient Boosting (3-day mortality): AUROC 0.8820 | Combining structured and unstructured data (physician notes) significantly improved mortality prediction accuracy. |
| **Tu et** | 2260 TBI | Single- | 42 features | LightGBM | In- | LightGBM | LightGBM with |
| **al., 2023** [21] | patients | center (Chi Mei Medical Center, Taiwan) | (vitals, GCS, pupil reflexes, muscle power, scores, treatments) from first day | , XGBoost, RF, LR; SHAP analysis | hospital mortality | (14 features): AUROC 0.914, Accuracy 0.878, Sens 0.806, Spec 0.886 | 14 first-day features showed excellent predictive accuracy for TBI mortality, outperforming APACHE II and SOFA. |
| **Wang et al., 2025** [2] | 69,190 ICU cases (general, older adults in WCHSU) | MIMIC-III, MIMIC-IV, WCHSU | Demographics, diagnoses, procedures, medications, 31 ICU observational indicators (first 48h) | CRISP framework (Causal Relationship Informed Superior Prediction): Transformer-based DL with causal graph and Counterfactual Minority Generation (CMG) | In-hospital mortality | MIMIC-III (CMG): AUROC 0.9480, AUPRC 0.7611. MIMIC-IV (CMG): AUROC 0.9171, AUPRC 0.6683. WCHSU (CMG): AUROC 0.9042, AUPRC 0.4771 | CRISP framework demonstrated stable and competitive performance across datasets, outperforming XGBoost. Causal inference integration shows promise. |

*(Note: AUROC = Area Under the Receiver Operating Characteristic curve; AUPRC = Area Under the Precision-Recall curve; Sens = Sensitivity; Spec = Specificity; LR = Logistic Regression; NB = Naïve Bayes; RF = Random Forest; SVM = Support Vector Machine; MLP = Multilayer Perceptron; GBT = Gradient Boosted Trees; NN = Neural Network; TD = Temporal Difference; LDA = Latent Dirichlet Allocation; FTIR = Fourier-Transform Infrared; FCBF = Fast Correlation-Based Filter; SHAP = SHapley Additive exPlanations; LIME = Local Interpretable Model-agnostic Explanations; APACHE = Acute Physiology and Chronic Health Evaluation; SAPS = Simplified Acute Physiology Score; SOFA = Sequential Organ Failure Assessment; LODS = Logistic Organ Dysfunction Score; GCS = Glasgow Coma Scale; CHF = Congestive Heart Failure; BUN = Blood Urea Nitrogen; PaO2 = Partial pressure of oxygen in arterial blood; PaCO2 = Partial pressure of carbon dioxide in arterial blood; AP-AKI = Acute Pancreatitis-Associated Acute Kidney Injury; TBI = Traumatic Brain Injury; AHRF = Acute*

*Hypoxemic Respiratory Failure; CMG = Counterfactual Minority Generation. Performance metrics are typically for the test/validation set unless otherwise specified. Confidence Intervals (CI) are included where provided in the source.)*